\pdfoutput=1
\documentclass[11pt]{article}

\usepackage{acl}
\usepackage{graphicx}
\usepackage[linesnumbered,ruled,vlined]{algorithm2e}
\usepackage{rotating}
\usepackage{color}
\usepackage{array}
\usepackage{caption}
\usepackage{multirow}
\usepackage{booktabs}
\usepackage{hhline}

\usepackage{times}
\usepackage{latexsym}

\usepackage[T1]{fontenc}

\usepackage[utf8]{inputenc}
\usepackage{xspace}

\usepackage{microtype}

\title{Semantic Compression for Word and Sentence Embeddings using Discrete Wavelet Transform}
\author{%
Rana Aref Salama\textsuperscript{1,2}, Abdou Youssef\textsuperscript{1}, and  Mona Diab\textsuperscript{3}\\
{\normalsize \textsuperscript{1} School of Engineering and Applied Science, George Washington University, USA}\\
{\normalsize \textsuperscript{2} Faculty of Computers and Artificial Intelligence, Cairo University, Egypt}\\
{ \normalsize \textsuperscript{3} Language Technologies Institute, Carnegie Mellon University, USA}
}

\begin{document}
\maketitle

\begin{abstract} 
Wavelet transforms, a powerful mathematical tool, have been widely used in different domains, including Signal and Image processing, to unravel intricate patterns, enhance data representation, and extract meaningful features from data. Tangible results from their application suggest that Wavelet transforms can be applied to NLP capturing a variety of linguistic and semantic properties.
In this paper, we empirically leverage the application of Discrete Wavelet Transforms (DWT) to word and sentence embeddings. We aim to showcase the capabilities of DWT in analyzing embedding representations at different levels of resolution and compressing them while maintaining their overall quality.
We assess the effectiveness of DWT embeddings on semantic similarity tasks to show how DWT can be used to consolidate important semantic information in an embedding vector. We show the efficacy of the proposed paradigm using different embedding models, including large language models, on downstream tasks. Our results show that DWT can reduce the dimensionality of embeddings by 50-93\% with almost no change in performance for semantic similarity tasks, while achieving superior accuracy in most downstream tasks. Our findings pave the way for applying DWT to improve NLP applications.
\end{abstract}

%============================================================================
\section{Introduction} \label{sec:1}
\vspace{-2mm}
Embedding models have evolved as a crucial part of any NLP application. Typically, they transform text, using different numerical analysis methods, into high-dimensional dense vectors that capture semantic and contextual aspects of the text for subsequent use by various tasks. As these models evolved, their complexity, dimensionality, and quality have all increased simultaneously, with a particular emphasis on quality and minimal attention to dimensionality. 
%Therefore, the investigation of the optimal embedding size for a model or task remains an unexplored challenge.
Typically generated embeddings are fixed in size for all tasks and are proportional to the model size, rendering their use in low resource settings a challenge. Accordingly, in this paper, we investigate the adaptation of DWT from Signal and Image processing to the field of NLP for the analysis and compression of word and sentence embeddings. DWT analyzes and reduces the size of the data by identifying redundant and important information in data. We posit that DWT has the potential to generate compressed embeddings capable of retaining contextual information and semantic relationships between words as well as among sentences.
% R:updated
Our key contributions: 1) Introduce DWT as an effective compression method for compressing embeddings with respect to an underlying task; 2) Propose a novel approach leveraging DWT for analyzing semantics in word and sentence embeddings; 3) Study the efficacy of DWT embeddings in capturing and retaining semantics by applying them to similarity and downstream tasks using various embeddings.%

\begin{figure}[t]
\centering
  \includegraphics[scale=0.5]{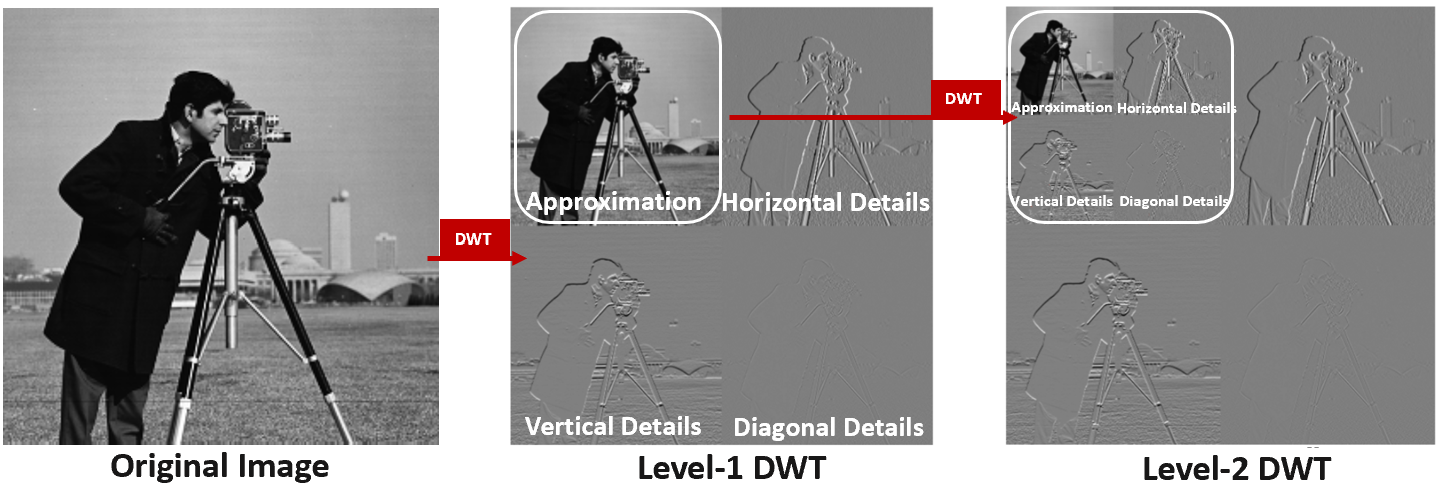}
  \caption {\label{camera} 
  \footnotesize Applying DWT to an image. Level-1 DWT transforms the image into approximation coefficients (that resemble the original image) and vertical, horizontal and diagonal detail coefficients. Level-2 DWT can be obtained recursively from Level-1 coefficients(multi-scale analysis).
  \vspace{-5mm}
  }
\end{figure}

\section{Motivation}
\label{sec:2}
\vspace{-1mm}
\vspace{-2mm}
The use of  spectral analysis methods in Image and Signal processing is driven by their ability to analyze data in the frequency domain, providing insights and revealing hidden patterns and dynamics not detectable in the spatial domain. Methods such as the Discrete Cosine Transform (DCT) and Fourier Transform (FT) have been commonly used. However, these methods offer global frequency representations that ignore the location of frequencies in the original domain and lack the ability to perform multi-resolution analysis. DWT effectively addresses these limitations by examining data change over time or position, i.e. frequency and time localization, and consequently captures varied information across different scales (multi-resolution analysis).
As a result, DWT achieved superior results in other fields. %in Image and Signal processing.
%=====================================
\begin{figure}[t]
\small
\centering
  \includegraphics[width=0.75\linewidth]{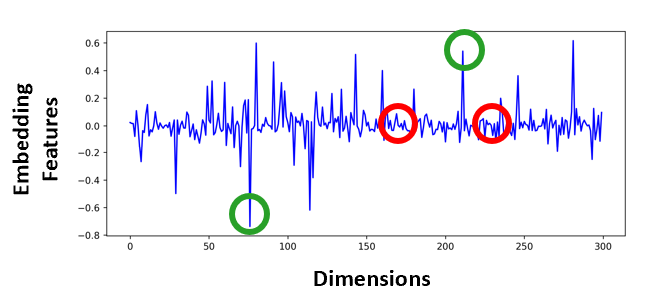} %\hfill
  \caption {\label{figword}  \footnotesize Graph of FastText embedding for the word 'work'(dim=300). Some features are similar in value (highlighted in red) and others exhibit significant spikes (highlighted in green).
  \vspace{-4mm}}
\end{figure}
%=====================================
DWT analyzes data based on their variation, filtering low-varying or highly correlated (low-frequency) and high-varying (high-frequency) components while preserving their spatial domain information. Figure\ref{camera} visualizes the result of applying DWT to an image resulting in  low-frequency coefficients (approximations) that capture the overall structure of the data, hence used for compression ~\citep{59,58}, while high-frequency coefficients (details) capture abrupt changes 
% AY
and motifs, 
which provide information about edges and contours, making them useful for edge detection ~\citep{11.5089125,60.ZHANG20091265} and noise filtering ~\citep{61.8404418}.
DWT can be further applied, recursively, to any coefficients of the first level to achieve a second level of the transform that contains more approximation and details coefficients. % as shown in Figure \ref{camera}.
In the realm of NLP, visualizing a word embedding vector, as shown in Figure\ref{figword}, we observe a signal-like structure with a few spikes indicating high variation in feature values, alongside features with minimal variation. This pattern suggests that applying spectral analysis to these vectors and numerical representations can reveal additional patterns or insights that may not be immediately apparent.
Few attempts for applying spectral methods to NLP including DCT~\citep{1.almarwani-etal-2019-efficient} and Higher-order Dynamic Mode Decomposition (HODMD)~\citep{30.kayal-tsatsaronis-2019-eigensent} were found successful. However, we are unaware of any application of applying DWT to word and sentence embedding other than our preliminary application of DWT to enhance DCT sentence embedding in \cite{dwt}. In this work, we posit that DWT can effectively compress embedding representations, and analyze them at different scales to understand and capture different linguistic patterns encoded in these embeddings. This analysis allows DWT to concentrate embedding energy with high compression ratios without significant loss of information. DWT allows for more efficient representation, less storage requirements and reduced computational complexity. Additionally, DWTs are scalable and can be applied to any embedding model and for any task. Accordingly, we believe that DWT, as a spectral transform, holds promise for NLP tasks.

%============================================================================
\vspace{-2mm}
\section{Related Work}
\label{sec:31}
\vspace{-2mm}
\subsection{Embedding Compression}
\vspace{-1mm}
Several studies have tackled embedding compression methods including code-book \citep{code}\citep{code2}, quantization \citep{quant}\citep{quant2}\cite{quant3} and factorization\cite{fact}. Other methods applied compression methods based on knowledge distillation \cite{distil}. Some other models consider compressing model parameters \cite{mdl_comp}, token embedding matrix \cite{direction}, or prune model weights \cite{prune}. 
Yet all these techniques compress only word representations, 
%AY
%considering embeddings as a vector of numbers, 
regardless of the semantics they convey.
Additionally, in the era of large language models (LLM), recent research \cite{dimensionality} argues that the dimensionality of the embedding representations, specifically in sentence embeddings, are sub-optimal and the same encoded information may be represented in smaller dimensions while achieving comparable performance.

\vspace{-2mm}
\subsection{Spectral Methods}
\vspace{-2mm}
The success of spectral methods in studying and analyzing embeddings has recently become evident in NLP. The application of DCT in sentence embedding has shown promising results\citep{1.almarwani-etal-2019-efficient}, 
%R:updated
yet it has only been applied to non-contextualized embeddings to generate sentence embeddings, not for compression. These DCT embeddings resulted in long sentence representations, corresponding to the number of coefficients considered from the transformation. We proposed using DWT to address this issue in \cite{dwt}, and results were promising.
%--
Similarly with HODMD ~\citep{30.kayal-tsatsaronis-2019-eigensent}, a sentence is represented as a signal with transitional properties captured in the frequency domain using uncorrelated coefficients to encode a sentence. Such models capture structural variation without losing on efficiency (comparable to averaging)~\citep{29.zhu2020sentence} and yet outperform more complex sentence embedding models~\citep{26.mikolov-etal-2018-advances}.
However, these models tend to analyze frequencies along similar word embedding dimensions, on a vertical level (inter-word embedding, aka 
across all words), accumulating a limited number of base frequency coefficients and \underline{dropping} the rest, in addition to ignoring their spatial domain position, i.e., ignoring  intra-word frequencies within individual word embeddings.

\vspace{-2mm}\section{Discrete Wavelet Transforms}
\label{sec:3}
\vspace{-2mm}
DWTs are mathematical functions that analyze data, \textit{f(t)}, using a window (a function) known as the {\em Mother Wavelet} (MW) $\psi(t)$, also called wavelets. This window moves over the data sequentially to examine the variations in the data with respect to this window (localized in time).  Within a given analysis window if surrounding data features exhibit minimal variation, they can be grouped together and represented using Approximation coefficients ($cA$). Conversely, if certain features vary significantly,  
%R:deleted diverging significantly,
they are represented using Detail coefficient ($cD$). To shift over the data, a MW translates and dilates $\psi_{a,b}(t)$ (as in equation (1)) using a shifting factor, \textit{b}, and scales with a scaling factor, \textit{a}, to capture different frequency variations at different data segments in time \textit{t}.
%R-deleted
%Low-scaled (diminished) MWs tend to capture abrupt changes in data, high frequencies, while %large-scaled (expanded) MWs capture low frequency variations.
%--
\vspace{-3mm}
\[\ \psi _{a,b}(t)=\frac{1}{\sqrt{a}}\psi(\frac{t-b}{a}) \ \ \ \ \ \ \ \  (1) \vspace{-2mm} \]
The resulting coefficients are equivalent to a pair of sub-band linear (convolutional) filters \citep{69}: one low-pass and one high-pass. The output of a filter pair is usually down-sampled by 2 so the combined output is of the same size (dimensionality) as the original input. This filtering + downsampling 
%R: deleted -is controlled by the MW function and-
can be cascaded on multiple levels recursively to analyze data at multiple levels of resolution. 
%R:deleted Figure~\ref{wav} shows the DWT of a 300-dimension word-embedding for two levels of DWT.
Note that DWT can be compared to Convolutional Neural Networks (CNNs)~\citep{74} in that they both use sliding-window filters and downsampling. 
%R:deleted (pooling)
The difference is that in CNNs, the filters are learned from the training data, while in DWT the filters are designed, not learned.
\\There are many families of MWs: Haar, Symmlets, Coiflets, and Daubechies, to name a few~\citep{20.1191457}. 
As a proof of concept in this paper, and in the interest of space, we will only be using and reporting on a subset of the MWs
%R:added
%AY
%and will only report 
that yield
the best results %achieved 
in our experiments.
%-
\footnote{For a more detailed explanation of Wavelet Transform theory,  refer to~\citep{73,20.1191457,68.brunton_kutz_2019}.}
%=============================================================================

\vspace{-2mm}
\section{Method} \label{sec:4}
\vspace{-2mm}
%R:updated
\begin{figure*}
\centering
  \includegraphics[]{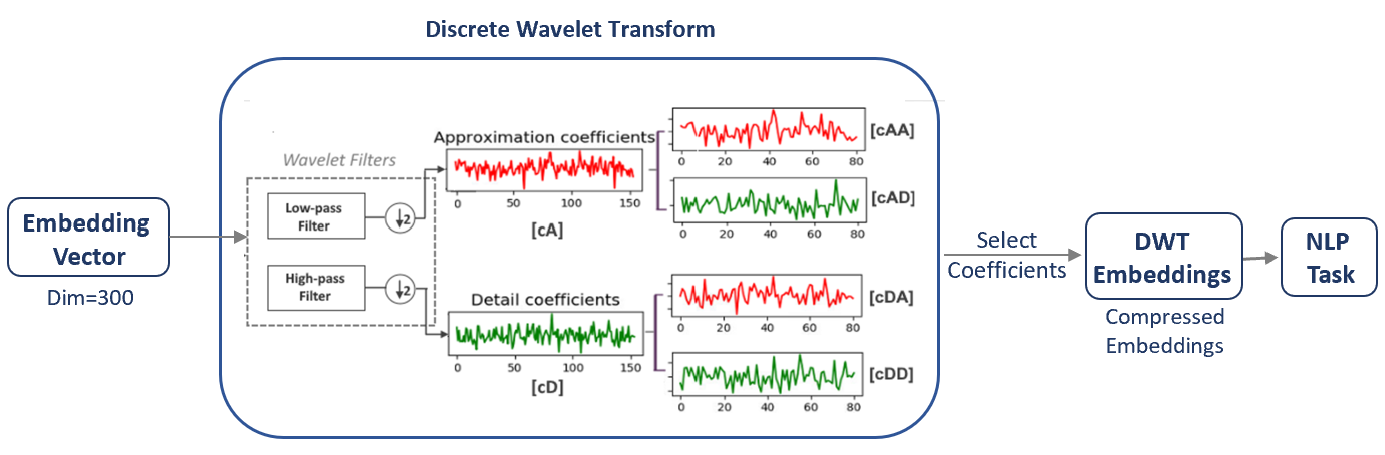} %\hfill
  \caption {\label{model_new} \footnotesize Transforming the embedding vector in Figure \ref{figword} using a 2-level DWT yielding cA and cD coefficients (dim=150) at Level-1. At Level-2, we get cAA and cAD from Level-1 approximations, and cDA and cDD from Level-1 details (dim=80).
  \vspace{-3mm}}
\end{figure*}
%--
%We explore applying DWT to word and sentence embeddings to study and analyze the features present in embedding representations. 
DWT filters features in embeddings by capturing how consecutive features in an embedding vector change, where small numerical difference or variation are converted into approximation coefficients. While features with large variations are converted into detail coefficients, and hence \textit{reflecting} the structure and behavior of the underlying information encoded in the vector.
%R:deleted 
%Adjacent features that have 
%. This process results in a more condensed compressed embedding at different scales that captures different levels of information.
%\paragraph{MW:}
%The first step for applying DWT is selecting a MW.\footnote{MWs are an essential component of DWT analysis, yet, for the ease and comprehension of this study, we will eliminate the details of the used MW for every experiment as this paper serves as a proof of concept.} In our experiments, we primarily utilize Symlets, Daubechies, and Coiflets MWs. The selection of a specific MW is empirically determined based on the best performance for a given task. %The findings regarding MW selection are concluded in Section\ref{sec:7} as an outcome of this study.
%\paragraph{Approach:}
%-
As shown in Figure\ref{model_new}, given a word or a sentence embedding $E_d$ of a \textit{d}-dimensional vector, we transform $E_d$ by applying DWT, $DWT(E)$, which decomposes the embedding vector into two 
%AY
%set 
vectors
of low-varying coefficients, cA, and high varying coefficients, cD, for one level of transformation. The dimension for each set of coefficients, $d'$, is reduced by 2, $d'=d/2$. The generated coefficients, cA and cD, can be further transformed for a second level, 
%of transformation 
generating 
%AY
%more 
new vectors of
approximation and detail coefficients, and downsampled by 2 again.
%AY: added
In fact, this recursive process can be repeated many times, say L times.
%2*(L=2), where 
L represents the number of DWT levels,
and %L 
is determined based on the required compression ratio and performance. 
%R:deleted We then use the generated coefficients as new embeddings, DWT embeddings. 
%R: added
For the selection of the MW used 
%R:deleted
in the transformation, we primarily use Symlets, Daubechies, and Coiflets wavelets across all experiments and the best performance per task is recorded. \footnote{In the interest of simplicity and clarity, we will omit the details of the specific MW used in each experiment, as this paper is intended as a proof of concept, 

and not for comparing different MWs.}
\\ \textbf{Coefficients Selection:}
In image processing, different sets of coefficients capture different aspects of an image. Similarly, we use different sets of coefficients as compact representations of the base embedding. We analyze the amount of information each set of coefficient encodes by studying their effectiveness in similarity and downstream tasks.
We consider the following possible selection mechanisms:
(a) \textit{Level-1 Coefficients:} Employ Level-1 coefficients, either the approximation (cA) or detail (cD) coefficients as the compressed representation with a size downsampled by 2.
(b) \textit{Higher-Level Coefficients:} Employ coefficients from higher 
levels of the DWT (i.e., Levels 2, 3 or later).
Specifically, we examine Level-2 approximation coefficients derived from Level-1 approximation and detail coefficients, namely cAA (approximation of approximation) and cDA (approximation of details)  
which are a quarter the size of the original embedding vector.
Additionally, for a higher compression ratio, we explore approximation coefficients from subsequent Level-3 (cAAA) and Level-4 (cAAAA),
Which are 1/8 and 1/16 the size of the original embedding vector, respectively.
Our emphasis lies on approximation for deeper compression, as they encapsulate an abstracted representation of the original embedding.
(c) \textit{Combined Coefficients:} In experiments where Level-1 coefficients do not achieve comparable performance, we incorporate coefficients from higher levels. 
Specifically, if Level-1 approximation doesn't sufficiently encode all relevant semantics, we further combine it with Level-2 approximation (derived from Level-1 detail), referred to as cDA. Similarly, for detail, we combine it with Level-2 detail (derived from Level-1 approximation), referred to as cAD. As illustrated in Figure \ref{model_new}, we enrich the red-colored coefficients from Level-1 (approximation) with the red-colored coefficients from Level-2 (obtained from Level-1 green coefficients). Thus, the combined embeddings are cA+cDA and similarly cD+cAD.
Note that combining Level-2 red-colored coefficients, cAA, with cA (Level-1 red-colored) is irrelevant since they contain redundant information. 
%>>
\vspace{-3mm}
\section{Evaluation} 
\vspace{-3mm}
To evaluate the proposed approach, we first investigate its efficacy in capturing and compressing semantics in semantic similarity tasks. We further evaluate their effectiveness in a number of downstream tasks to evaluate their efficacy extrinsically. We consider different embeddings, baselines, experimental setup and tasks as follows. 
\\\textbf{Embeddings:} %consider moving with every section
%To demonstrate the capability of DWT 
We experiment with different embeddings; %including: 
Pre-trained Language Models as BERT \citep{44.bert}, GPT \citep{gpt}, SBERT \cite{45.wang2020SBERTwk} and RoBERTa \cite{RoBERTa} sentence embeddings. We also include non-contextualized embeddings from our previous work \cite{dwt} for completeness.
We use GloVe \cite{85} and FastText \cite{26.mikolov-etal-2018-advances} with various dimensions (50, 100, 200, 300).
%=================================
\begin{table}
\centering
\small
 \setlength{\tabcolsep}{0.7\tabcolsep}
\begin{tabular}{l|cccc} 
\hline
\textbf{Word Embedding} & \textbf{Dim} & \textbf{SimLex} & \textbf{WS353} & \textbf{MEN} \\ 
\hline
GloVe\textsubscript{100} & 100 & 12.22 & 46.96 & 57.73 \\
GloVe\textsubscript{50}  & 50 & 9.82 & 42.17 & 53.05 \\ 
\hline
GloVe+DWTcD & 50 & 11.50 & \textbf{\textcolor{red}{50.19}} & \textbf{\textcolor{red}{58.96}} \\
GloVe+DWTcA & 50 & \textbf{\textcolor{red}{13.48} } & 44.08 & 57.21 \\ 
\hline
\multicolumn{1}{l}{} & \multicolumn{1}{l}{} & \multicolumn{1}{l}{} & \multicolumn{1}{l}{} & \multicolumn{1}{l}{} \\ 
\hline
GloVe\textsubscript{200}  & 200 & 13.03 & 48.00 & 59.42 \\
GloVe\textsubscript{100}  & 100 & 12.22 & 46.96 & 57.73 \\ 
\hline
GloVe+DWTcD & 100 & 11.50 & \textcolor{red}{\textbf{50.19}} & 58.96 \\
Glove+DWTcA & 100 & \textbf{\textcolor{red}{20.79}} & \textcolor{red}{\textbf{50.19}} & \textbf{\textcolor{red}{62.00}} \\ 
\hline
\multicolumn{1}{l}{} & \multicolumn{1}{l}{} & \multicolumn{1}{l}{} & \multicolumn{1}{l}{} & \multicolumn{1}{l}{} \\ 
\hline
FastText & 300 & 50.30 & \textbf{ 79.13} & \textbf{83.36 } \\
PCA & 150 & 27.01 & 52.22 & 63.31 \\ 
\hline
FastText+DWT\textsubscript{cD} & 150 & \textbf{\textcolor{red}{50.32}} & 74.18 & 79.91 \\
FastTex+DWT\textsubscript{cA} & 150 & 49.03 & \textcolor{red}{75.44} & \textcolor{red}{80.96} \\
FastText+DWT\textsubscript{cD+cAD} & 225 & \textbf{\textcolor{red}{50.32}} & \textcolor{red}{78.34} & \textcolor{red}{82.59} \\ 
FastText+DWT\textsubscript{cA+cDA} & 225 & 49.05 & 75.56 & \textcolor{red}{82.96} \\
\hline
\end{tabular}
\caption{\label{sim-non}  \footnotesize Spearman Rank Order Correlation (SPC) results on SimLex-999, WS353 and MEN datasets; using GloVe-Twitter27B embeddings compared to Level-1 DWT coefficients. Baseline includes base GloVe embeddings of similar size in addition to GloVe with 50\% less dimensions. DWT\textsubscript{cD} and DWT\textsubscript{cA} correspond to the embeddings yielded at Level-1 DWT transform. Level-2 DWT\textsubscript{cD+cAD} and DWT\textsubscript{cA+cDA} coefficients from Level-1 coefficients(dim=150) concatenated with Level-2 (dim=75) Best results are in bold and best results per experimental condition are in red.
\vspace{-6mm}
}
\end{table}
\\\textbf{Baselines:}
For every experiment, we use different baselines, in addition to the base embeddings, to explore the capabilities of DWT.
Our baselines include: (1) Base embeddings (in all experiments); (2) Other dimensionality reduction methods to assess their effectiveness in comparison to DWT: PCA for dimensionality reduction\cite{81.pca} and DCT for compression\citep{dct} in some of our experiments.
\\\textbf{Experimental Setup:} 
%\vspace{-2mm}
For all sentence embeddings experiments, we use the SentEval toolkit~\citep{28.DBLP:journals/corr/abs-1803-05449} for evaluation. For all downstream tasks, we leverage multi-layer perceptron (MLP) classifiers based on the default setup outlined in SentEval.\footnote{The source code is publicly available on GitHub at https://github.com/engranas/DWT-Semantic-Compression}
%=================================
\vspace{-3mm}
\subsection{ Semantic Similarity Evaluation} 
\vspace{-1mm}
Word and sentence similarity tasks have become the de-facto method for semantic evaluation \cite{jrank}. Semantic Similarity involves measuring the degree of relatedness and similarity between pairs of words or sentences compared against human judgments or similarity scores assigned by human annotators.

\begin{table}
\centering
\footnotesize
\begin{tabular}{l|cccc} 
\hline
 & \textbf{\textbf{Dim}} & \textbf{\textbf{SimLex}} & \textbf{\textbf{WS353}} & \textbf{MEN} \\ 
\hline
BERTbase & 768 & 60.75 & 28.00 & \textbf{59.55} \\ 
\hline
BERT+DWTcD & 383 & 60.31 & \textbf{\textcolor{red}{28.41}} & 59.19 \\
BERT+DWTcA & 383 & \textbf{\textcolor{red}{60.90}} & 28.25 & \textcolor{red}{59.31} \\ 
\hline
BERTLarge & 1024 & 69.72 & 44.00 & 62.18 \\ 
\hline
BERT+DWTcD & 512 & 69.65 & \textbf{\textcolor{red}{45.05}} & \textbf{\textcolor{red}{62.68} } \\
BERT+DWTcA & 512 & \textbf{\textcolor{red}{69.95}} & 43.31 & 61.09 \\ 
\hline
\multicolumn{1}{l}{} & \multicolumn{1}{l}{} & \multicolumn{1}{l}{} & \multicolumn{1}{l}{} & \multicolumn{1}{l}{} \\ 
\hline
GPTBase & 1536 & \textbf{50.00} & \textbf{64.89} & 73.00 \\ 
\hline
GPT+DWTcD & 768 & \textcolor{red}{49.95} & 63.88 & \textbf{\textcolor{red}{73.34}} \\
GPT+DWTcA & 768 & 49.37 & \textcolor{red}{64.58} & 71.82 \\ 
\hline
GPTLarge & 3072 & 56.60 & 72.31 & 78.35 \\ 
\hline
GPT+DWTcD & 1535 & 56.23 & 72.03 & 77.57 \\
GPT+DWTcA & 1535 & \textcolor{red}{\textbf{56.93}} & 71.93 & 77.29 \\
\hline
\end{tabular}
\caption{\label{sim-con}  \footnotesize Similar experiment as Table \ref{sim-non} using BERT and GPT base and large word embeddings with base embeddings as the baseline.
\vspace{-3mm}}
\end{table}
\vspace{-2mm}
\subsubsection{Word Similarity Evaluation}

We start by evaluating DWT embeddings for word similarity,
%AY
%. In this evaluation, we consider 
using
the following datasets: SimLex-999~\citep{23.DBLP:journals/corr/HillRK14}, MEN~\citep{25.10.5555/2655713.2655714} and WS353~\citep{24}.
\\In our initial experiment, we utilize DWT Level-1 approximation (cA) and detail (cD) coefficients as DWT embeddings for the word semantic similarity task. In this experiment we assess:
(1) GloVe embeddings with dimensions 100 and 200. Baselines are: the base GloVe embeddings of the same size, alongside the base GloVe embeddings originally reduced in dimensions by 50\%.
%<<R:deleted
%as embedding representations to evaluate . Specifically both the  coefficients serve as DWT embeddings, each reduced by 50\% from the original embedding size. We consider a number of parameterized and non-parameterized embeddings including (1) GloVe embeddings, with varying dimensions (100, 200, 300) for a comprehensive evaluation, baselines are: the original embeddings (referred to as GloVe\textsubscript{1}) and, GloVe\textsubscript{2}, representing the original GloVe embeddings with dimensions reduced by 50\% to compare their efficacy with compressed embeddings.
%>>
(2) FastText embeddings with dimension 300. Baselines are: the base FastText embeddings, 
%in addition to 
and PCA reduced embedding with size 150 dimensions. 
(3) BERT and GPT models, where for each model we consider two variants, base and large models, in order to conduct a thorough evaluation with original embeddings as baselines. We found DWT embeddings to consistently outperform the baselines for GloVe embeddings, with dimensions 100 and 200, as depicted in Table~\ref{sim-non}. 
DWT embeddings efficiently compress the semantics encoded in the original embeddings, surpassing both baselines with a 50\% reduction in dimensionality and in some cases surpassing the performance of embeddings that are four times larger as in SimLex and WS353, where DWT embeddings of size 50 outperforms GloVe embeddings of size 200.  This empirically shows that DWT embeddings not only serve  as a dimensionality reduction technique but also adeptly capture the semantics encoded in an embedding vector and effectively compress them in DWT coefficients.
Additionally, DWT embeddings significantly outperform PCA-reduced embeddings, demonstrating superior performance with  >20\% improvement.
Additionally, in the case of contextualized embeddings, DWT embeddings exhibit comparable performance to all baselines for all models. This suggests that the large dimensionality used for these models may not be significant for encoding words. It can also be noted that the DWT BERT embeddings are more comparable to the baseline than DWT GPT embeddings; this suggests that GPT models are packing more information and semantics than the BERT model for these datasets. This empirically shows that DWT reveals new aspects of embeddings that were not evident in the original domain.
Nevertheless, for FastText, while DWT yields similar performance to the original embeddings for the SimLex dataset, yet for WS353 and MEN datasets Level-1 DWT embeddings failed to fully capture the encoded semantics present in the original embeddings. This discrepancy suggests that FastText embeddings contain richer semantics compared to GloVe embeddings, a conclusion supported by the similarity results achieved with the original embeddings.
%===============================
\begin{table*}
\footnotesize
\centering
\resizebox{\textwidth}{!}{
\begin{tabular}{|l|c|c|c|} 
\hline
\textbf{Word} & \textbf{BERT}(dim=768) & \textbf{BERT+DWT\textsubscript{cA}}(dim=383) & \textbf{BERT+DWT\textsubscript{cD}}(dim=383) \\ 
\hline
happy & happy, sad, pleased, smiling, thrilled & happy, smiling, excited, pleased, glad &happy, sad, joy, pleased, thrilled\\ 
\hline
sea & sea, gulf, marine, desert, underwater & sea, marine, desert, gulf, island & sea, gulf, underwater, beach, fish\\ 
\hline
playing & playing, riding, practicing, throwing, creating& playing, creating, riding, writing, coloring & playing, riding, fighting, throwing, plays \\
\hline
\end{tabular}}
\caption{\label{5nn} \vspace{-2mm} \footnotesize 5-nearest cosine similar words using BERT embeddings.}
\vspace{-3mm}
\end{table*}
%
%========================
%<<R:Table deleted
%\begin{table}[h] 
%\centering
%\small
%\begin{tabular}{l|cccc} 
%\hline
% & \textbf{Dim} & \textbf{SimLex} & \textbf{WS353} & \textbf{MEN} \\ 
%\hline
%FastText & 300 & \textbf{50.30} & \textbf{79.13} & \textbf{83.36} \\ 
%\hline
%DWT\textsubscript{cD+cAD} & 225 & \textbf{\textcolor{red}{50.32}} & \textcolor{red {78.34} & \textcolor{red}{82.59} \\ 
%\hline
%DWT\textsubscript{cA+cDA} & 225 & 49.05 & 75.56 & \textcolor{red}{82.96} \\
%\hline
%\end{tabular}
%\caption{\label{simF}%Word Similarity Evaluation using FastText Embeddings on SimLex-999, WS353 and MEN datasets.
%Similar experiment as Table \ref{sim-non} using cD+cAD and cA+cDA coefficients from Level-1 coefficients(dim=150) concatenated with Level-2 (dim=75).%Best results are in bold and best results per experimental condition are in red.
%}
%\end{table}
%=========================
\\We further use the combined coefficients; cA+cDA and cA+cDA. As shown in Table~\ref{sim-non}, the combined DWT embeddings at a compression of 25\% dimensionality augmenting the embeddings with additional semantics, resulting in a performance comparable to the baseline with a slight reduction approx. 1\% in performance for WS353 and MEN datasets. Conversely, for SimLex datasets, the addition of coefficients from subsequent layers (cD+cAD) did not improve the performance and cD coefficients seems to have effectively encoded the semantics for all word embeddings.
Nevertheless, it is crucial to emphasize that the selection of DWT coefficients 
%AY doesn't 
depends not only  %solely 
on the type of embeddings but also on the particular task in consideration. As a result, we utilize DWT embeddings generated from FastText for another semantic task, Concept Categorization, to further elaborate on the adaptability of DWT for different tasks. Concept Categorization groups words in different categories based on semantic clusters~\citep{54.baroni-etal-2014-dont}. 
%That proved effective in downstream NLP tasks~\citep{22.wang_wang_chen_wang_kuo_2019}.
For this evaluation we use the datasets: AP~\citep{55.article}, BM~\citep{57.bm.5555/2387636.2387658} and BLESS datasets~\citep{56.bless.inproceedings}. We use the base embeddings as the baseline.
%==========================
\begin{table}
\centering
\footnotesize
\begin{tabular}{lccccc} 
\hline
  & \textbf{Dim} & \textbf{AP} & \textbf{BM} & \textbf{BLESS}   \\ 
\hline
FastText& 300  & \textbf{0.70}    & \textbf{0.47}  & \textbf{0.86}  \\ 
\hline
FastText+DWT\textsubscript{cD}& 150    & \textbf{\textcolor{red}{ 0.70}}   & 0.46  & \textbf{\textcolor{red}{0.87}}  \\ 
\hline
FastText+DWT\textsubscript{cA} & 150    & \textbf{\textcolor{red}{0.70}}& \textbf{\textcolor{red}{ 0.49}}  & 0.82  \\
\hline
\end{tabular}
\caption{ \label{cg} \footnotesize Results for Concept Categorization task for 3 standard datasets: AP, BM and BLESS using using Level-1 cA and cD coefficients. %Best results are in bold and best results per experimental condition are in red
\vspace{-3mm}}
\end{table}
As illustrated in Table~\ref{cg}, cA and cD embeddings demonstrate comparable or even superior results compared to the baseline, despite 50\% dimensionality reduction. This illustrates the effectiveness of DWT embeddings in encapsulating essential semantics from the original embedding for the given task.
%==========================
\begin{table*}
\centering
\scriptsize
\begin{tabular}{l|lcccc||l|lcccc}
\hline
\textbf{Model}             & \textbf{Dim}  & \textbf{STS12} & \textbf{STS16} & \textbf{STSB}  & \textbf{SICKR} & \textbf{Model}                                     & \textbf{Dim}  & \textbf{STS12} & \textbf{STS16} & \textbf{STSB}  & \textbf{SICKR} \\ \hline
 SBERT\textsubscript{Base}  & 768  & 74.09 & 84.08 & 85.35 & \textbf{80.69} &   RoBERTa\textsubscript{Base} & 768    & 69.02 & 83.39 & 81.89 & 80.57 \\
   SBERT+DCT   & 384  & 74.14 & 84.2  & 75.57 & 53.05    & 
   RoBERTa+DCT   &  384    & 68.64 & 82.87 & 65.01 & 39.14 \\ 
         SBERT+DCT   & 192  &   72.69    & 83.40      &   74.99    &     51.55  &                
         RoBERTa+DCT   & 192      & 67.97 & 82.53 & 67.14 & 36.30  \\ \hline
         SBERT+DWT\textsubscript{cD}    & 384  & 73.57 & 83.83 & 85.83 & 80.07 &                RoBERTa+DWT\textsubscript{cD}    &   384   & \textcolor{red}{\textbf{ 69.32}} & \textcolor{red}{\textbf{ 83.57}} & 82.40  & \textcolor{red}{\textbf{ 80.79}} \\
          SBERT+DWT\textsubscript{cA} & 384  & \textcolor{red}{\textbf{74.26}} & \textcolor{red}{\textbf{84.27}} &\textcolor{red}{\textbf{ 85.98}} & \textcolor{red}{80.55} &               RoBERTa+DWT\textsubscript{cA}   &  384    & 68.88 & 82.93 & 82.48 & 80.71 \\
           SBERT+DWT\textsubscript{cAA} & 192  & 73.72 & 83.9  & 85.77 & 80.02 &                RoBERTa+DWT\textsubscript{cAA}   &  192    & 68.17 & 82.63 & 82.63 & 80.71 \\
          SBERT+DWT\textsubscript{cDA} & 192  & 73.21 & 83.90  & 85.70  & 79.56 &                RoBERTa+DWT\textsubscript{cDA}   & 192     & 68.63 & 82.91 & \textcolor{red}{\textbf{ 83.60}}  & 80.31 \\ \hline \hline
    SBERT\textsubscript{Large} & 1024 & 67.97 & 81.69 & 78.26 & 80.24 &   
   RoBERTa\textsubscript{Large} & 1024 & 65.07 & 75.70  & 74.68 & 79.86 \\
        SBERT+DCT   & 512  & 67.83    & 75.66 & 62.99 & 41.92 &                
        RoBERTa+DCT   & 512  & 64.79 & 75.07 & 44.00 & 37.23  \\
       SBERT+DCT   & 256  &     65.41  & 73.65      & 61.55      &  40.25     &                
       RoBERTa+DCT   & 256  & 63.05      &74.12       &  30.52     & 28.23      \\ \hline
     SBERT+DWT\textsubscript{cD}  & 512  & 67.83 & 81.49 & 80.97 & \textcolor{red}{\textbf{ 80.95}} &                 
     RoBERTa+DWT\textsubscript{cD}   & 512  & \textcolor{red}{\textbf{65.07}} & \textcolor{red}{\textbf{75.70}}  &76.76 & 79.65 \\
     SBERT+DWT\textsubscript{cA} & 512  & \textcolor{red}{\textbf{ 67.97}} &\textcolor{red}{\textbf{  81.69}} & 80.15 & 80.44 &                
      RoBERTa+DWT\textsubscript{cA}& 512  & 64.97 & 75.67 & 75.72 & \textcolor{red}{79.68} \\
     SBERT+DWT\textsubscript{cAA}& 256  & 67.96 & 81.34 & \textcolor{red}{\textbf{ 82.42}} & 79.69 &                
     RoBERTa+DWT\textsubscript{cAA}   & 256  & 64.94 & 75.5  &  \textcolor{red}{\textbf{77.14}} & 78.71 \\
     SBERT+DWT\textsubscript{cDA}   & 256  & 67.57 & 81.15 & 82.33 & 79.94 &              RoBERTa+DWT\textsubscript{cDA} & 256  & 64.87 & 75.14 & \textcolor{red}{\textbf{ 77.14}} & 79.03 \\ \hline
\end{tabular}
\caption{\label{pretrained} \footnotesize Results on the STS benchmark, Spearman’s correlation is reported. Baseline represents the original and DCT reduced embeddings for  SBERT and RoBERTa models. The best overall results are shown in bold. Best results per condition are shown in red.
\vspace{-4mm}}
\end{table*}
%==========================
To conduct a qualitative analysis of DWT embeddings, 
%AY: the \ref{5nn} wasn't working, because the \label{5nn} was not included within the caption. So I added it into the caption of that table. 
Table \ref{5nn} displays the five nearest neighbors (determined by cosine similarity) for randomly chosen words utilizing BERT word embeddings. As shown, approximation coefficients capture more relevant words like 'excited' and 'glad' for the word 'happy'. Also, 'island' for 'sea', and 'writing' and 'coloring' for the word 'playing'. Detail coefficients capture more appropriate words such as 'joy' for 'happy', 'beach' and 'fish' for 'sea', and 'fighting' for 'playing'.
Additionally, it's apparent that certain relevant words, which share similar meanings or contexts, have closer similarity, such as "creating" for "playing" and "smiling" for "happy".
%*******************************************
\vspace{-1mm}
\subsubsection{Sentence Semantic Similarity}
\vspace{-1mm}
The Semantic Textual Similarity (STS) Task is a common benchmark used for evaluating the performance of semantic models. In this setting, we examine the application of DWT to contextualized sentence embeddings on STS tasks 2012 and 2016~\citep{sts12,sts16}, STS benchmark (STSB)~\citep{stsb} and SICK-Relatedness~\citep{sickr}. (See Appendix for more STS tasks results.) 
We evaluate DWT embeddings on contextualized models as they are becoming dominant for sentence embedding~\citep{sentllm}. We consider SBERT base and large models \cite{45.wang2020SBERTwk}, RoBERTa base and large models \cite{RoBERTa}.\footnote{Models available in https://huggingface.co/sentence-transformers; SBERT-base-nli-v2, SBERT-Large-nli-v2, nli-roberta-base and nli-roberta-large.}
In this experiment, we will further explore the effectiveness of DWT in capturing relevant features and semantics within pretrained language models, illustrating that the extended size of dimensionality of these embeddings is not optimal for semantic representation, which we found consistent with the recent results and findings in \cite{dimensionality} without the need for any further training or fine-tuning of the model.
We consider Level-1 coefficients, cA and cD, with 50\% reduction in dimensions. We also consider Level-2 coefficients, cAA and cDA, with 75\% reduction in dimension. For the baselines we consider: (1) The base embeddings, (2) DCT~\citep{dct} as another spectral model used for lossy compression. 
We use DCT to compress the original SBERT and RoBERTa embeddings by applying the DCT transform to these vectors and selecting the first n coefficients, where n is equivalent to 50\% or 25\% of the original embedding size.\footnote{In this context, DCT is not applied in the same way as proposed by \citep{1.almarwani-etal-2019-efficient}, where the transform is applied across all words in a sentence to encode the sentence. In our context we apply it within a word vector.}
As shown in Table \ref{pretrained}, DWT consistently surpasses DCT across all tasks, particularly in STSB and SICKR, where the efficacy of DCT embeddings reflects significantly lower performance. Although DCT was capable of compressing embeddings and maintaining comparable performance in STS12 and STS16, %However, 
DCT compresses by discarding residual frequency coefficients leading to the loss of important details and a subsequent decline in performance in STSB and SICKR. 
DWT performs similarly to the base embeddings, with at most a 0.5\% decrease in performance on a few tasks for all models using Level-1 coefficients in SBERT and RoBERTa. This indicates that Level-1 coefficients effectively capture relevant semantics, condensing them into fewer dimensions. Notably, for SBERT, the approximation coefficients yield better performance, while for RoBERTa, the detail coefficients perform better. This empirically demonstrates that SBERT and RoBERTa capture different types of features: SBERT captures more relational semantic features, whereas RoBERTa focuses on distinct semantic features.
%<<R:deleted
%DWT embedding effectively achieves similar performance to the baseline except for a few tasks; using the base models in STS12, yet it is still comparable. Also for STS14 except when using RoBERTa Large with Level-2 coefficients, the condensed representation beats the baseline margin. 
%>>
Additionally, DWT reveals some characteristics of these embeddings: large models tend to contain more redundant features. As a result, DWT can effectively compress them to 75\% less dimensions while maintaining comparable performance across all tasks.
More interestingly, for the STSB task using SBERT and RoBERTa, DWT outperforms the baseline in a manner proportional to the model size. Specifically, in the large models, cAA coefficients improve performance by 4\% in SBERT and by 2.5\% in RoBERTa. This shows that the cAA coefficients contain more dense semantic information suggesting that the original embeddings had highly correlated dimensions with redundant features that proved to be irrelevant when filtered out using DWT. 
%This can be further proved by how cAA and cDA, equally outperforms the baseline in STSB task by more than 2\% in performance. This
%However, for STS16 and SICKR, DWT embeddings outperfoms, while for STSB, RoBERTa model, Base and Large, outperforms the baseline with a ratio compatible with the model size. 
Generally, DWT embeddings prove to have comparable performance or better in most tasks, at 50\% reduction in embedding size. At a 75\% reduction in size, the performance is still comparable and never drops more than 2\%, and for some tasks the performance is significantly better. 
In most cases cA represents a good approximation for the original representation.
%>> end of section update

%----------------------------------------------------------------------------

\begin{table*}
\centering
\resizebox{\textwidth}{!}{
\tiny
\begin{tabular}{lc|ccccc|c|c|c|c} 
\hline
                 & \multicolumn{1}{l|}{} & \multicolumn{5}{c|}{\textbf{Sentiment Analysis} }                                                                                                                   & \multicolumn{1}{c|}{\textbf{Inference} } &\textbf{Paraphrase}                               & \textbf{SUBJ}& \textbf{TREC} \\ 
\hline
\textbf{Embedding} 
                 & \textbf{Dim}          & \textbf{MR}                      & \textbf{CR}                      & \textbf{SST2}           & \textbf{SST5}                    & \textbf{MPQA}                    & \textbf{SICK-E}                         & \textbf{MRPC}          &                        &                        \\ 
\hline
%PCA & 128 & 72.76 & 79.93 & 76.93 & 38.02& 76.03& - &62.13&84.33&68.00\\
%\hline
RoBERTa\textsubscript{Base} & 768& 85.32	&90.46	&92.31	&52.17	&89.17	&80.01&	70.72&	\textbf{94.71}	&\textbf{92.40}        \\ 
\hline
DWT\textsubscript{cA} & 384& 85.28	&91.20	&91.82&	53.57	&\textcolor{red}{\textbf{89.70}}	&79.16	&73.62&	94.21	&\textcolor{red}{90.60}    \\ 
DWT\textsubscript{cD}& 384 & \textcolor{red}{\textbf{85.34}}&	90.97 &\textcolor{red}{\textbf{92.53}}&	\textcolor{red}{\textbf{53.71}}&	89.38&	79.58&	72.23&	\textcolor{red}{94.38}&	90.41
\\ 

DWT\textsubscript{cAA} & 192 & 84.59	&\textcolor{red}{\textbf{91.36}}	&90.94	&53.03	&89.18	&\textcolor{red}{\textbf{80.31}}	&\textcolor{red}{\textbf{73.97}}	&93.23	&84.82 \\ 

DWT\textsubscript{cDA}  & 192  &84.78	&91.13	&91.43	&52.94&	89.27	&79.93	&73.57&	93.32	&82.60\\
\hline 
\hline
RoBERTa\textsubscript{Large}        & 1024  & 
85.01&	91.18&	91.38	&50.95&	90.13	&80.68&	76.17	&92.00&	85.84\\ 
\hline
DWT\textsubscript{cA} & 512 &\textcolor{red}{\textbf{85.97}}&	91.26&	91.60	&52.22&	90.62&	\textcolor{red}{\textbf{82.04}}	&77.22&	92.21	&87.20            
\\ 
DWT\textsubscript{cD}& 512 & 85.57&	\textcolor{red}{\textbf{91.44}}&	91.71&	\textcolor{red}{\textbf{55.34}}&	90.48&	81.63&	77.16&	\textcolor{red}{\textbf{92.43}}&	\textcolor{red}{\textbf{87.60}} \\ 

DWT\textsubscript{cAA} & 256 & 
85.39&	91.29	&91.65&	53.85&	\textcolor{red}{\textbf{90.72}}	&81.33&	\textcolor{red}{\textbf{77.39}}	&91.80	&85.20
\\ 

DWT\textsubscript{cDA}  & 256  &
85.40&	91.07&	\textcolor{red}{\textbf{91.93}}&	53.39	&90.51&	80.96&	76.93&	92.08&	85.80
\\
\hline
\end{tabular}
}
\caption{\label{extrinsic} \footnotesize Best Classification accuracy results on various classification tasks for Level-1 approxmation and details coefficients; cA and cD, and Level-2 DWT coefficients; cAA and cDA. The Baseline is the original RoBERTa for Base and Large models.  The best overall results are shown in bold. Best results per condition are shown in red.
\vspace{-4mm}
} 
\end{table*}
\vspace{-2mm}
\subsection{ Downstream Tasks}
\vspace{-2mm}
\label{down}
To further evaluate our model extrinsically, we explore applying DWT embeddings to the following downstream tasks: sentiment classification on Movie Reviews (MR),  Stanford Sentiment Treebank (SST2, SST5)~\citep{32.pang-lee-2004-sentimental}, product review (CR)~\citep{33.Hu2004MiningAS}, subjectivity classification (SUBJ)~\citep{34.pang-lee-2004-sentimental}, opinion polarity classification (MPQA), question type classification (TREC)~\citep{36.10.1145/345508.345577}, paraphrase identification (MRPC)~\citep{37.dolan-etal-2004-unsupervised}, and entailment classification on the SICK dataset (SICK-E)~\citep{12.inbook}.
%<<R:added
Due space limitation, we will only consider RoBERTa base and large embeddings for these experiments as a proof of concept, to show the efficacy of DWT. % given the space limitation.
%>>
Table~\ref{extrinsic} shows the results with base embeddings as the baseline. As shown, DWT %embeddings 
effectively compresses the base embeddings, outperforming the baselines in all tasks. Level-1 coefficients, cD, achieve better performance by 4.5\% in SST5 task, 2\% in TREC task, while cA outperforms in MR and SICK-E. Using Level-2 coefficients, cAA, further improves over the baseline in MPQA and MRPC, while cDA outperforms in SST2. 
Generally, DWT embedding surpassed the baselines with all coefficients for all tasks except for CR and TREC where the Level-2 coefficients is  comparable. This demonstrates the efficacy of DWT embeddings for downstream tasks.
In conclusion, we find that DWT presents an effective balance between efficiency (compactness) and accuracy, serving as an efficient data-size reduction method that condenses embeddings with relevant features, leading to enhanced performance.
%<<R:updated
\\ \textbf{Multiple Levels of DWT} We further investigate the effectiveness of applying multiple levels of DWT transformation to RoBERTa large embeddings. We extend our analysis up to 4 levels of transformation, thereby achieving compression of up to 93\%. We apply DWT recursively to the approximation coefficients from each level, resulting in coefficients of sizes 512, 256, 128, and 64 respectively.
%===============================
\begin{figure}[t]
\tiny
\centering
 \includegraphics[scale=0.2]{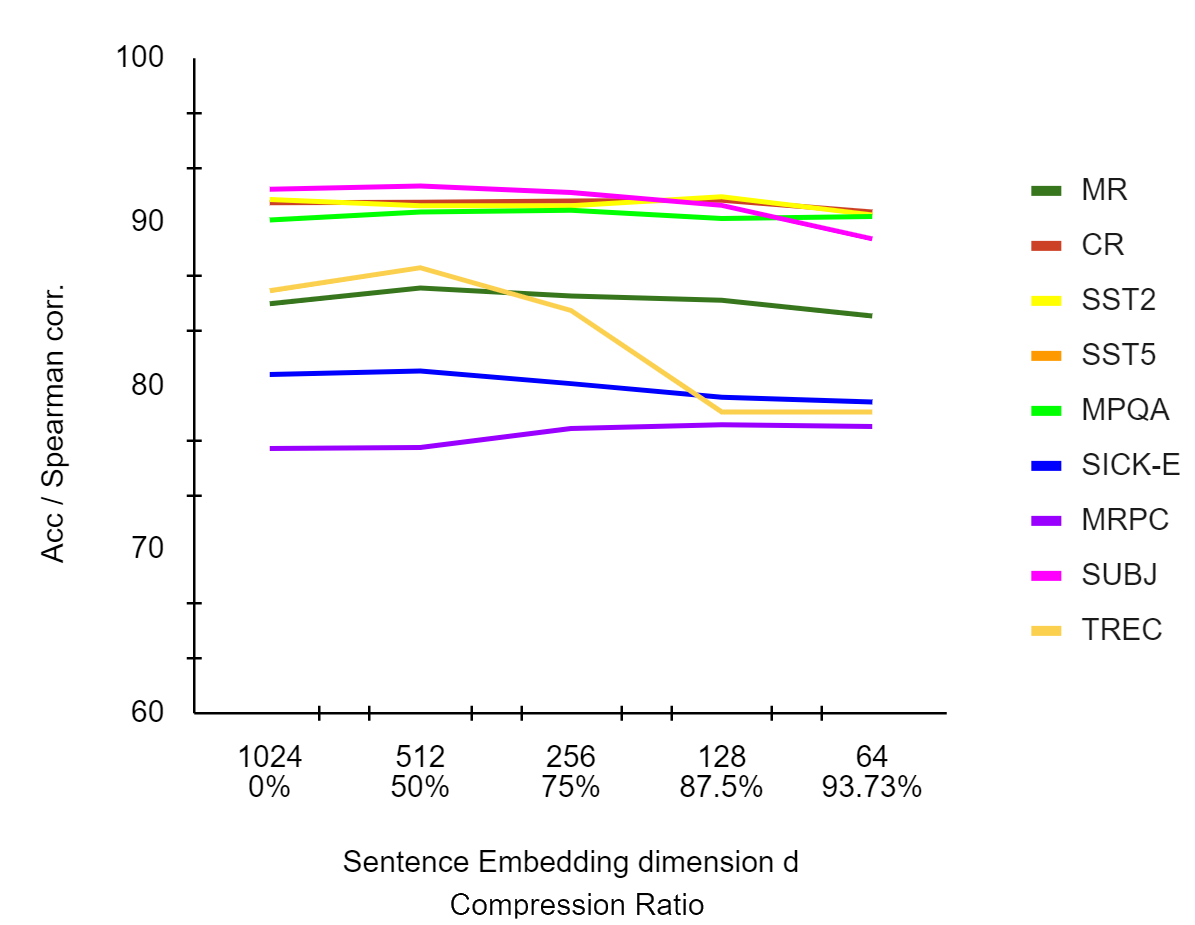}
    \caption{ \footnotesize Accuracy results for 4-Levels of DWT for approximation coefficients for downstream tasks using RoBERTa Large embeddings
      \vspace{-5mm}
    } 
    \label{4level}
    
\end{figure}
%===============================
The results\footnote{Detailed results for this experiment can be found in the Appendix.}in Figure \ref{4level} show that for dimensions of size 128 and 64, corresponding to Level-3 and Level-4 coefficients, with a reduction in dimensionality of 87\%-94\%, respectively. DWT embeddings outperform the baseline for all tasks by 0.09-1.5\% except for SICK-E, SUBJ, and TREC, which experience degradation by 1.3\%, 1\%, and 7.4\%, respectively, although still outperforming the baseline in higher levels. This decline in performance tends to correlate with an increased compression ratio indicating that the minimized dimension size is not sufficient to fully encode the necessary semantic information, resulting in a tolerable performance decline. 
On the other hand, other tasks like MRPC outperform the baseline by 1.5\% with just 64 dimensions, indicating that DWT effectively consolidates relevant features into much fewer dimensions, while the extended dimension size contains redundant features.
These findings align with those of \cite{dimensionality}, where their model, despite being trained using a two-step method, consistently degrades performance for all tasks by 2\%-9\% in dimension sizes of 64. In contrast, our model exhibits comparability to the baseline in most tasks and experiences degradation by 2\%-3.5\% in MR, SICK-E, and SUBJ, and by 7.4\% in TREC.
\vspace{-2mm}
\section{Conclusion and Discussion}
\label{sec:7}
%\vspace{-1mm}
\vspace{-2mm}
In this paper, we explored the effectiveness of applying DWT to word and sentence embeddings to \textit{selectively} reduce embeddings. Our experiments illustrate the potential of DWT to enhance space and computational efficiency without decline in
performance, reducing embedding size by 50-75\%. DWT reveals new insights about embeddings, and exposes hidden patterns and semantic information that were not apparent in the base embedding space. Additionally, the generated DWT embeddings postulate that different sets of coefficients capture different semantic aspects of an embedding. 
We conclude our study outcomes in the following points:
%<<R:updated
\\\textbf{Correlation and Redundancy:} While it has not been proven that features within the same embedding are correlated, the comparable performance achieved by DWT suggests a correlation between the dimensions of an embedding. This challenges the previous assumption that these dimensions are uncorrelated. If no correlation existed, DWT would hardly achieve comparable results to the performance of base embeddings. Our results also prove that embeddings from complex models, such as Pre-trained Language Models, contain redundant features and hence compressing them with 75\% reduction in dimensionality achieves a comparable performance to the base embeddings with no more than 2\% degradation.
%>>
\\\textbf{Dimensionality Reduction Techniques:} Although the general idea of DWT appears to be similar to other dimensionality reduction methods in the context of decomposition and compression, %yet, 
its overall properties and methodology are different~\citep{67}. 
%Consequently, 
DWT outperforms other compression techniques such as PCA and DCT. 
DWT compresses the data using localized sub-band frequency components capturing both low and high features-variation. DWT stands out as a 
% non-linear %AY: it is linear filtering
method that can be applied universally to diverse datasets and models.
%AY: applying to different datasets is not scalability, but universality, which you already mentioned
%, providing scalability across different types of data. % While 
PCA reduces data to a low dimension space, and de-correlates the data, 
%AY: added
but is much slower because it needs to compute 
%using a linear combination of 
the eigenvectors of the original data~\citep{pca}. As for DCT, despite being a frequency analysis method, %yet DCT 
it 
tend to capture important frequencies indiscriminately, lacking localization and discarding other frequencies and accordingly degrading performance. 
\\ \textbf{Mother Wavelets:}
MW is a component of DWT that controls feature filtering, with each MW corresponding to a unique pair of filters. %While we 
We did not explore specific MW applications for each task due to space constraints, but limited ourselves to a predetermined number of MW families such that we maintain experimental consistency while empirically investigating the effectiveness of DWT. Our experiments indicate that Coiflets wavelets generally perform well across tasks. It is important to note that MW filtering is controlled by a scaling factor, which adjusts the level of variation considered in an embedding. For example, a Coiflet MW with scale 4 focuses on nearby features and their fine variations, whereas scale 17 emphasizes broader approximations, omitting fine details. 
%AY: deleted and added
Since the choice of the best MW varies from base embedding to base embedding, 
%Additionally, MW output varies depending on the embeddings, thus a pre-processing step may be required to identify the best MW and coefficients. However, 
the selection and optimization of MW requires further study beyond this paper's scope. 
%and is a pertinent research inquiry in the domains of image and signal processing. 
As a result, we leave further exploration of MW details and selection for future research. 
%>>
%<<R:added
\\ \textbf{Coefficients Selection: }
Much as in image processing, different coefficients tend to capture distinct aspects of the data. Therefore, it is essential to select the set of coefficients that are most suitable for the base embedding being used in a given task. Some tasks may benefit more from approximations over details, while others may see improved performance with nuanced information, and yet others may use both equally. However, it can be concluded that if minimizing the embedding size is a significant consideration, utilizing only Level-1 coefficients (cA or cD) with a 50\% reduction in size results in comparable performance to the baselines across nearly all tasks. Moreover, approximation coefficients consistently maintain comparable performance across multiple levels of DWT, as demonstrated in the 4-level analysis, with the exception of TREC. This consistency suggests that these coefficients retain relevant information about the original embeddings. Nevertheless, a detailed study and analysis of coefficient selection fall beyond the scope of this paper and are planned for future work.
\\ \textbf{DWT for Compression: }
DWT offers a powerful method for compressing embeddings by exploiting correlations and reducing redundancy. The performance of DWT in compressing embeddings strongly suggests that approximation coefficients can be effectively utilized for compression, akin to their role in image compression, where they typically capture essential features of the data. In the compression process, approximation coefficients not only reduce the embedding dimensionality but also reduce noise while retaining relevant features. By reducing %or eliminating 
noise, the performance of the model improves as it casts down on the error rate thus improving the accuracy of these results. 
%AY: deleted the next sentence because it is redundant with the previous sentence
%This improvement in performance is attributed to compression's ability to reduce noise in the representation, thereby enhancing the accuracy of the outcome.
%Additionally, 
Selecting the appropriate compression level for a given task is a hyperparameter that 
%is contingent upon 
depends on
various factors, including space complexity, resources allocated for a particular task and the trade-off between effectiveness and compression ratio.  
\\ \textbf{Computational Complexity:}
The DWT transformation is implemented through convolution and down-sampling (filtering) operations, which are typically linear with respect to the size of the input data. The overall computational complexity is often expressed in terms of the dimension of the embedding, $d$, so for a single level of transformation, the complexity is $O(d)$. However, the recursive nature of multi-level transformations can lead to increased computations, depending on the number of levels, $L$, resulting in an overall complexity of $O(L×d)$. Yet, DWT facilitates efficient dimensionality reduction, significantly reducing memory usage and increasing throughput by shrinking the size of embeddings while preserving essential features.
\\ \textbf{DWT Efficacy: }Based on our evaluations, the application of DWT in the context of NLP exhibits significant potential for efficiently modeling word and sentence embeddings. Our findings demonstrate that DWT coefficients have the ability to capture various aspects of the data, with the approximation coefficients serving as a general approximation of an embedding, akin to their behavior in image analysis. Furthermore, the details coefficients excel at capturing semantic nuances within the embeddings. Notably, DWT embeddings reveal new aspects and characteristics that were previously unexplored, such as establishing correlations between dimensions of a word embedding. 
Additionally, DWT can adapt to any embedding: words, sentences or documents of any length in the same manner as mentioned in this paper.
Overall, the application of DWT holds promise, and we anticipate its effectiveness in other NLP applications.
%AY: why do you want a new page here? 
%\newpage
\section{Limitations}
In this paper, our focus is to thoroughly and empirically investigate the effectiveness of applying Discrete Wavelet Transform (DWT) to word and sentence embeddings, with a primary emphasis on analysis and compression. However, DWT transformations are mainly based on the selection of Mother Wavelets (MWs), which we did not specify for every experiment due to space constraints. We employed Coiflets, Symlets, Haar, and Daubechies as MWs, with the recorded best overall results. Choosing the optimal MW for DWT is a common research question in Image and Signal processing, and we defer the investigation into the selection of the best MW to future work.
The results presented in this paper serve as a proof of concept, indicating the potential for other MWs to further enhance performance. 
Additionally, our results were generated using publicly available models that undergo regular updates; thus, discrepancies may exist between our results and those published by the model owners.
\section{Ethical Considerations}\label{sec:ethical_considerations}
As we propose a novel method for applying DWT to NLP embedding, this section is divided into the following two parts.
\subsection{ Dataset}
\paragraph{Intellectual Properties and Privacy Rights}
We make use of publicly-available data for all experiments with no modification or update for fair comparison.
\subsection{NLP Application}
\paragraph{Code Availability}
When used as intended, applying the models described in this paper can save people much time. Our code will be publically available to ensure reproduciblity of results
\paragraph{Code Reusability}
To run our experiments, we sometimes used public Github repositories as intended by their authors, without any modification. All such code was appropriately referenced. 
\paragraph{Environmental Cost}
The experiments described in the paper makes no use of GPUs. We used CPU for our experiments. The experiments run in several hours. Several dozen experiments were run due to parameter search, and future work should experiment with distilled models for more light-weight training. We note that while our work required extensive experiments to draw sound conclusions, future work will be able to draw on these insights and need not run as many large-scale comparisons. Models in production may be trained once for use using the most promising settings.

\bibliography{anthology,custom}
\bibliographystyle{acl_natbib}

\appendix 
\newpage
\section{Appendix}
%In this Appendix we provide few detailed experiments to showcase our results and help comprehend  

\subsection{Sentence Semantic Similarity}

\begin{figure}[ht]
\centering
 \includegraphics[scale=0.38]{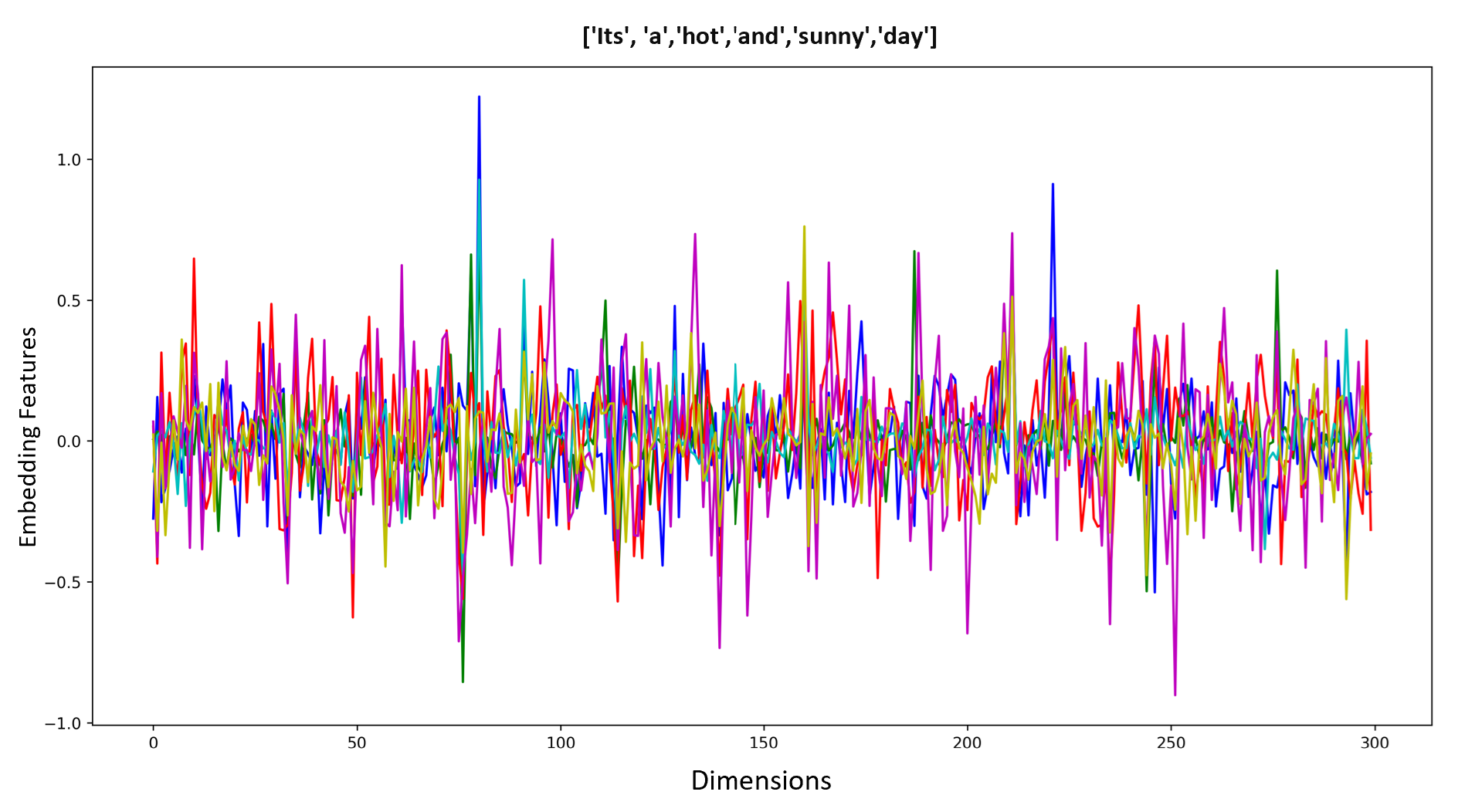}
  \caption{ \label{avg} Similarity Matrix between 2 sentences using BERT embeddings of dimension 768, and Level-1 cA and cD of the transformed BERT embedding with dimension of 384.}
\end{figure}

\paragraph{Visualizing Embeddings of Words in a Sentence}
By sketching out all the words in a sentence like "it's a hot and sunny day," as shown in Figure \ref{avg}, we observe that the energy of all word embeddings overlaps within an average sub-band. This provides a compelling explanation for why word averaging effectively represents a sentence, offering empirical support for averaging as a method for sentence embedding. This suggests that further spectral analysis of these embedding representations is promising and likely to achieve effective results, similar to those in image and signal processing.

\paragraph{Qualitative Analysis}
%==================================
\begin{figure*}[ht]
\centering
 \includegraphics[scale=0.7]{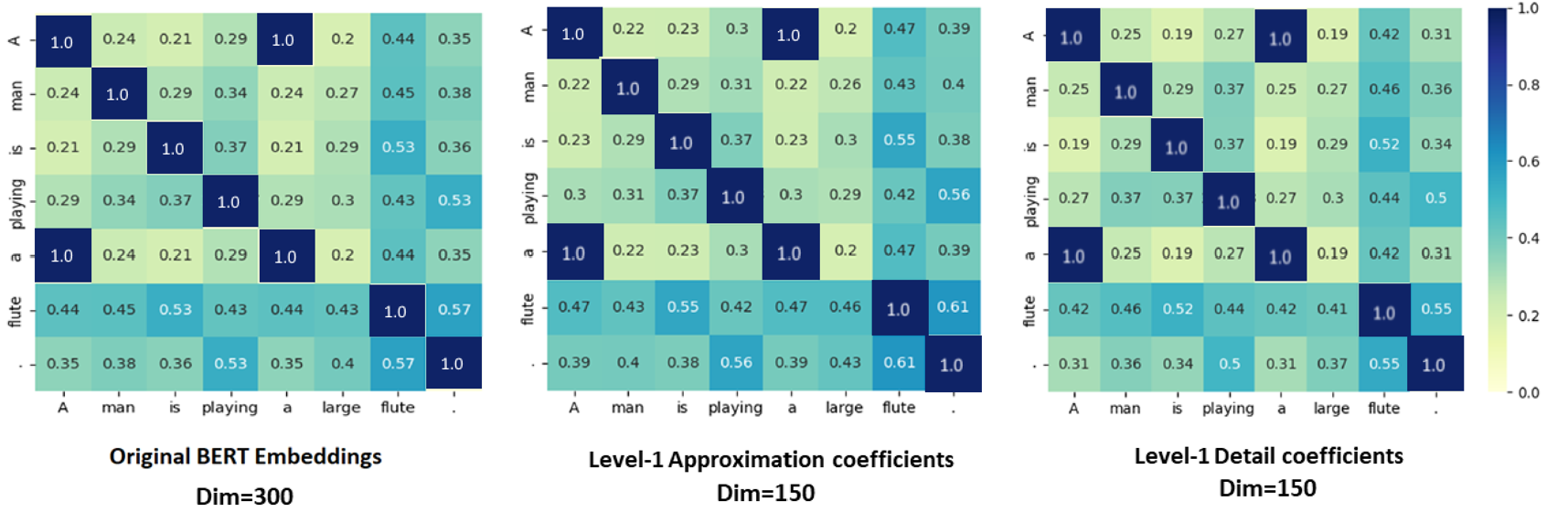}
  \caption{ \label{bert} Similarity Matrix between 2 sentences using BERT embeddings of dimension 768, and Level-1 cA and cD of the transformed BERT embedding with dimension of 384.
  }
\end{figure*}
%==================================
To further explore the efficacy of DWT through Sentence Similarity Tasks, we initially demonstrates the word similarity matrix between two randomly selected sentences from the STSB dataset using BERT\cite{44.bert} embeddings with a dimension size of 768 as opposed to their Level-1 DWT cA and cD coefficients, with a dimension size down-sampled by 2, i.e. dimension size is 384. By comparing the overall distribution of similarity shown in Figure \ref{bert}, we observe that it remains largely coherent between the original embeddings and the DWT embeddings with 50\% reduction in dimensions.

%==================================
\begin{table*}
\centering
\small
\begin{tabular}{llcccccccc}
\hline
\textbf{Model}         & \textbf{} & \textbf{Dim} & \textbf{STS12} & \textbf{STS13} & \textbf{STS14} & \textbf{STS15} & \textbf{STS16} & \textbf{STSB} & \textbf{SICKR} \\ \hline
InferSent & Baseline  & 4096 &
                    50.05&	45.64&	57.40&	62.21	&59.44	& \textbf{67.19}&	\textbf{81.95}\\
                    \hline
                    & DWT\textsubscript{cD} & 2053 & 52.93	& \textbf{\textcolor{red}{48.05}}	& \textbf{\textcolor{red}{58.23}}	&63.90	&61.63	& \textcolor{red}{66.27}	&\textcolor{red}{81.06} \\
                       & DWT\textsubscript{cA}  & 2053  & 48.16	&44.32	&56.13	&60.78	&58.04	&64.91	&80.57   \\
                       & DWT\textsubscript{cAA}       & 1030          & 
                       46.09	&42.64	&54.36	&59.06	&56.83&	61.55	&78.76
                       \\
                       & DWT\textsubscript{cDA}       & 1030          & 
                       \textbf{\textcolor{red}{53.10}}&	47.87&	58.13&	 \textbf{\textcolor{red}{63.94}}&	 \textbf{\textcolor{red}{61.80}}&	65.73&	79.31
                       \\ \hline

\end{tabular}
\caption{\label{infersent}Results on the STS  benchmark, Spearman’s correlation is reported. Baseline represents the original embedding and corresponding performance for InferSent model. The best overall results are shown in bold. Best results per condition are shown in red.}
\end{table*}
%==================================

\paragraph{Extended Intrinsic Evaluation:}
\paragraph{1- InferSent Embeddings}
We additionally consider InferSent sentence embedding model\cite{infersent} for this experiment as an example for 
a parameterized sentence embedding model. We set the original embeddings as the baseline.
As shown in Table\ref{infersent}, Level-1 cD coefficients outperform the baselines for all tasks by 1.5-3\% better performance except for STSB ad SICKR which have comparable results. Level-2 cDA coefficients exceed the performance in STS12,STS15 and STS16 at a dimension reduction by 75\% showing (1) that DWT condensed more relevant semantics for these task in subsequent levels of transformation and (2) that the nuance in the features represented in the original embeddings were more relevant for these tasks in the InferSent representation context, having cD outperforms cA coefficients. Still, cA coefficients also beat the baseline.

%==================================
\begin{table}
\tiny
\centering
\begin{tabular}{llccccccc}
\hline
\textbf{Model}         & \textbf{} & \textbf{Dim}  & \textbf{STS13} & \textbf{STS14} & \textbf{STS15} \\ \hline

SBERT\textsubscript{Base} & Baseline  & 768                   & 83.56          & \textbf{90.73}          & \textbf{88.03}                  \\
                       & SBERT+DWT\textsubscript{cD}       & 384                   & 83.06          & \textcolor{red}{90.55}          & 87.56               \\
                       & SBERT+DWT\textsubscript{cA}        & 384                  & \textbf{\textcolor{red}{83.60}}          & 90.44          & \textcolor{red}{87.77}            \\
                       & SBERT+DWT\textsubscript{cAA}       & 192               & 82.05          & 90.11           & 86.99           \\
                       & SBERT+DWT\textsubscript{cDA}       & 192                  & 81.58          & 90.11          & 86.40                   \\ \hline
SBERT\textsubscript{Large} & Baseline  & 1024          	&78.79&	\textbf{79.41}&	\textbf{82.20}         \\
                       & SBERT+DWT\textsubscript{cD}        & 512         &	 \textcolor{red}{\textbf{78.86}}	&\textcolor{red}{79.36}	&82.00\\
                       & SBERT+DWT\textsubscript{cA}        & 512          &78.16&	79.30&	\textcolor{red}{82.02}       \\
                       &SBERT+DWT\textsubscript{cAA}   &256    &	76.68&	78.93&	81.59\\
                       & SBERT+DWT\textsubscript{cDA}       & 256         &	78.26&	78.95&	81.22	
                       \\ \hline
RoBERTa\textsubscript{Base}                & Baseline  &      768       &	80.17&	\textbf{80.47}&	84.04 \\
                       & RoBERTa+DWT\textsubscript{cD}       &      384      &	\textcolor{red}{\textbf{80.34}}&	\textcolor{red}{80.38}&	\textcolor{red}{\textbf{84.64}}
                       \\
                       & RoBERTa+DWT\textsubscript{cA}        &   384           & 
                      	79.38&	80.35&	83.68
                       \\
                       & RoBERTa+DWT\textsubscript{cAA}       &         192     & 
                      	78.76&	79.81&	82.16
                       \\
                       &RoBERTa+DWT\textsubscript{cDA}    &            192  & 
                      	79.58&	79.95&	82.95
\\ \hline
RoBERTa\textsubscript{Large}                & Baseline  &      1024                 & 70.34          & 72.41          & 77.60              \\
                       & RoBERTa+DWT\textsubscript{cD}        &     512                  & 70.00            & 72.30          & 77.51              \\
                       & RoBERTa+DWT\textsubscript{cA}        &   512                 &  \textbf{\textcolor{red}{70.34}}         & 72.58         & 77.58               \\
                       & RoBERTa+DWT\textsubscript{cAA}       &         192            & 69.68         &  \textbf{\textcolor{red}{72.72}}         & \textbf{\textcolor{red}{77.61}}              \\
                       & RoBERTa+DWT\textsubscript{cDA}       &            192         & 69.52         & 72.44         & 76.73                  \\ \hline

%BERT                   & Baseline  &              & 0.700          & 0.7446         & 0.7594         & 0.8339         & 0.8131         & 0.8242        & 0.7990         \\
%                       & cD        &              & 0.6931         & 0.753          & 0.7527         & 0.8304         & 0.8103         & 0.8221        & 0.7922         \\
%                       & cA        &              & 0.7005         & 0.7332         & 0.7570         & 0.8248         & 0.8046         & 0.8188        & 0.7936         \\
%                      & cAA       &              & 0.6891         & 0.7182         & 0.7473         & 0.8047         & 0.7953         & 0.8108        & 0.7796         \\
%                       & cDA       &              & 0.6856         & 0.7348         & 0.7477         & 0.8170         & 0.8016         & 0.8145        & 0.7804         \\ \hline
\end{tabular}
\caption{\label{extrasim} \footnotesize Results on the STS13-STS15  benchmark, Spearman’s correlation is reported. Baseline represents the original embedding and corresponding performance for SBERT and RoBERTa models. The best overall results are shown in bold. Best results per condition are shown in red.}
\end{table}
%==================================

\paragraph{2- Semantic Similarity Tasks (2013-2015) }
Subsequently, we present the results for applying DWT on STS tasks 2013-2015~\citep{sts13,sts14,sts15} using Pre-trained Language Model embeddings, SBERT and RoBERTa, for a detailed study for the performance of DWT on more semantic similarity tasks. Table \ref{extrasim} shows the result of applying DWT embeddings using SBERT and RoBERTa models. As shown, DWT outperforms the baselines for STS13 and is very comparable to STS14 and STS15. 

%=================================
\subsection{4-Level DWT Embedding Results}
In this section we represent the detailed results for Figure\ref{4level} for 4-Levels of DWT tasks applied to downstream tasks included in Section\ref{down} using RoBERTa Large embeddings. Table\ref{4leveldata} shows the results for Level-1 approximation coefficients, cA, Level-2 approximation coefficients cAA, Level-3 approximation coefficients, cAAA and Level-4 coefficients, cAAAA. The results show the efficacy of DWT approximation coefficients to maintain relevant information despite decreasing the size of the embedding by more than 90\%.
%=================================
\begin{table*}[ht]
\vspace{-50em}
\centering
\setlength{\tabcolsep}{0.85\tabcolsep}
\begin{tabular}{l|lrrrrrrrrr}
\hline
\textbf{Model} & \multicolumn{1}{l}{\textbf{Dim}} & \multicolumn{1}{l}{\textbf{MR}}       & \multicolumn{1}{l}{\textbf{CR}}       & \multicolumn{1}{l}{\textbf{SST2}}     & \multicolumn{1}{l}{\textbf{SST5}}     & \multicolumn{1}{l}{\textbf{MPQA}}     & \multicolumn{1}{l}{\textbf{SICK-E}}   & \multicolumn{1}{l}{\textbf{MRPC}}     & \multicolumn{1}{l}{\textbf{SUBJ}}     & \multicolumn{1}{l}{\textbf{TREC}}    \\ \hline
RoBERTa\textsubscript{Large}  & \multicolumn{1}{l}{1024}         & 85.01                                 & 91.18                                 & 91.38                                 & 50.95                                 & 90.13                                 & 80.68                                 & 76.17                                 & 92.00                                    & 85.80                                 \\ \hline
DWT\textsubscript{cA}          & 512                              & {\color[HTML]{FE0000} \textbf{85.97}} & 91.21                                 & 90.99                                 & 51.04                                 & 90.62                                 & 80.64                                 & 77.23 & {\color[HTML]{FE0000} \textbf{92.21}} & {\color[HTML]{FE0000} \textbf{87.20}} \\
DWT\textsubscript{cAA}        & 256                              & 85.49                                 & {\color[HTML]{FE0000} \textbf{91.29}} & 90.99                                 & {\color[HTML]{FE0000} \textbf{53.85}} & {\color[HTML]{FE0000} \textbf{90.72}} & {\color[HTML]{FE0000} \textbf{81.33}} & 77.39                                & 91.80                                  & 84.00                                \\
DWT\textsubscript{cAAA}       & 128                              & 85.22                                 & 91.34                                 & {\color[HTML]{FE0000} \textbf{91.54}} & 52.35                                 & 90.22                                 & 79.03                                 & 77.62                                 & 91.02                                 & 82.40                                \\
DWT\textsubscript{cAAAA}     & 64                               & 84.26                                 & 90.62                                 & 90.44                                 & 50.05                                 & 90.34                                 & 76.74                                 &{\color[HTML]{FE0000} \textbf{77.51}}                              & 88.97                                 & 78.40                                 \\ \hline
\end{tabular}
\caption{\label{4leveldata}
Best Classification accuracy results on various classification tasks for Level-1 approxmation; cA, Level-2 DWT coefficients; cAA, Level-3 DWT coefficients; cAAA and Level-4 DWT coefficients; cAAAA. The Baseline is the original RoBERTa Large model. The best overall results are shown in bold. Best results per condition are shown in red}
\end{table*}

\end{document}